\pdfoutput=1
\PassOptionsToPackage{table}{xcolor}
\documentclass[11pt]{article}
\usepackage[final]{coling}
\usepackage{times}
\usepackage{latexsym}
\usepackage[T1]{fontenc}
\usepackage[utf8]{inputenc}

\usepackage{microtype}
\usepackage{inconsolata}
\usepackage{graphicx}
\usepackage{tipa}
\usepackage{CJKutf8}
\usepackage{amsmath}
\usepackage{amsfonts}
\usepackage{algorithm}
\usepackage{algpseudocode}
\usepackage{amssymb}
\usepackage{caption,subcaption}
\usepackage{array}
\usepackage{pifont}
\usepackage{geometry}
\usepackage{booktabs}
\usepackage{tabularx}
\usepackage{enumitem}
\usepackage{hyperref}

\newcommand\blfootnote[1]{%
  \begingroup
  \renewcommand\thefootnote{}\footnote{#1}%
  \addtocounter{footnote}{-1}%
  \endgroup
}

\title{Swift Cross-Dataset Pruning: \\ Enhancing Fine-Tuning Efficiency in Natural Language Understanding}

\author{
 \textbf{Binh-Nguyen Nguyen\textsuperscript{1,3}} \and
 \textbf{Yang He\textsuperscript{1,2}\thanks{Corresponding Author}}
\\
 \textsuperscript{1}CFAR, Agency for Science, Technology and Research, Singapore
 \\
 \textsuperscript{2}IHPC, Agency for Science, Technology and Research, Singapore
 \\
 \textsuperscript{3}VNU University of Engineering and Technology, Hanoi, Vietnam
  \\
 \texttt{21020526@vnu.edu.vn}, \texttt{he\_yang@cfar.a-star.edu.sg}
}

\begin{document}
\maketitle
\begin{abstract}
Dataset pruning aims to select a subset of a dataset for efficient model training. While data efficiency in natural language processing has primarily focused on within-corpus scenarios during model pre-training, efficient dataset pruning for task-specific fine-tuning across diverse datasets remains challenging due to variability in dataset sizes, data distributions, class imbalance and label spaces. Current cross-dataset pruning techniques for fine-tuning often rely on computationally expensive sample ranking processes, typically requiring full dataset training or reference models. We address this gap by proposing \textbf{Swift Cross-Dataset Pruning (SCDP)}. Specifically, our approach uses TF-IDF embeddings with geometric median to rapidly evaluate sample importance.  We then apply dataset size-adaptive pruning to ensure diversity: for smaller datasets, we retain samples far from the geometric median, while for larger ones, we employ distance-based stratified pruning. Experimental results on six diverse datasets demonstrate the effectiveness of our method, spanning various tasks and scales while significantly reducing computational resources. Source code is available at: \url{https://github.com/he-y/NLP-Dataset-Pruning}.
\end{abstract}

\section{Introduction}
\blfootnote{This work was completed during Binh-Nguyen Nguyen's internship at CFAR, A*STAR.}

\begin{figure}[t]
    \centering
    \includegraphics[width=\linewidth]{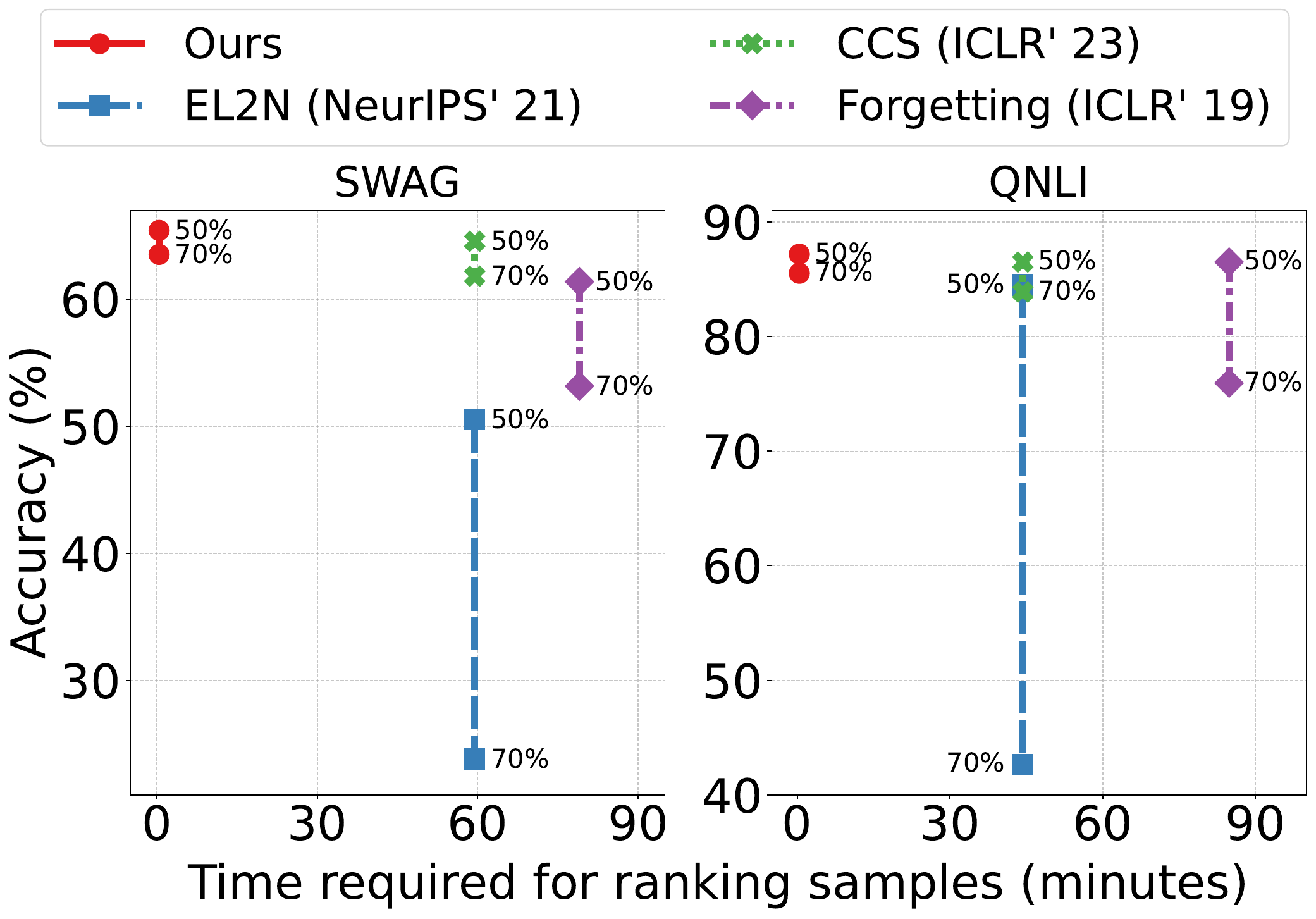}
    \caption{Accuracy and time required for ranking samples for SWAG, QNLI datasets with our proposed method, EL2N, CCS, Forgetting at 50\% pruning rate and 70\% pruning rate. Our method is significantly more time-efficient and yields higher accuracy.}
    \label{fig:fig1}
\end{figure}

Deep learning progress has been fueled by massive datasets \cite{tan2024data,gadre2024datacomp}, but managing and training on such data poses computational and storage challenges \cite{yangdataset}. Dataset pruning, or coreset selection, aims to identify a subset that achieves comparable model performance to the full dataset \cite{mirzasoleiman2020coresets,killamsetty2021glister}, reducing training and storage costs while maintaining model effectiveness \cite{huang2021novel,xia2022moderate}. 

This challenge is particularly evident in language model (LM) training, which involves two distinct scenarios: pre-training and fine-tuning, each requiring different data handling approaches. 
Data for pre-training, like those used for BERT, comprise large-scale, unlabeled, and diverse corpora like BookCorpus and English Wikipedia, collectively containing over 3,300 million words \cite{devlin-etal-2019-bert}. These corpora aim to facilitate the learning of broad language representations. 
In contrast, datasets for fine-tuning for downstream tasks, such as SWAG (113,000 examples) \cite{zellers2018swag}, are smaller, labeled, and task-specific, designed to evaluate targeted abilities such as commonsense reasoning. Although dataset efficiency techniques such as language filtering, quality assessment, and deduplication \cite{albalak2024a, longpre2024pretrainer} are proposed for large-scale pre-training corpora, they are not suitable for fine-tuning.

\begin{figure*}[!t]
    \centering
    \includegraphics[width=\linewidth]{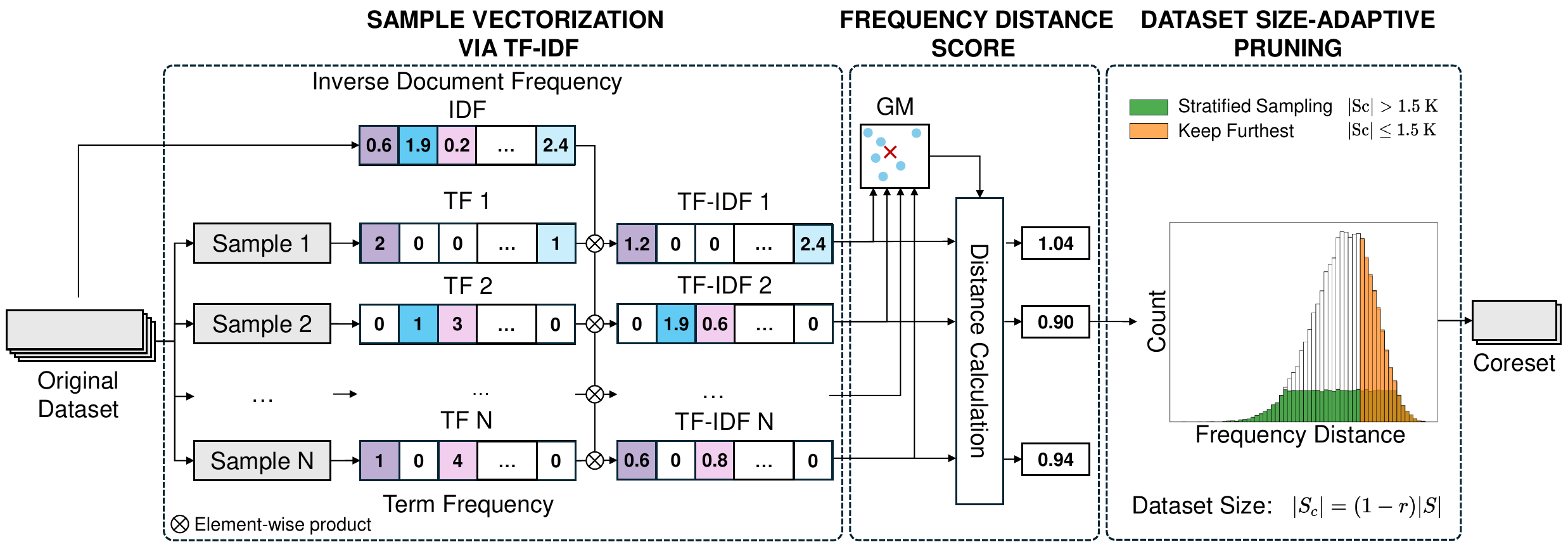}
    \caption{Overview of the proposed method. We introduce the Frequency Distance (FD) score, in which we leverage TF-IDF embeddings combined with geometric median calculations to swiftly assess sample importance. We propose dataset size-adaptive pruning to enhance adaptability in cross-dataset setting.}
    \label{fig:fig2}
\end{figure*}

While single-dataset fine-tuning benefits from a narrow target distribution \cite{albalak2024a}, establishing general dataset pruning rules for cross-dataset fine-tuning remains challenging due to the diversity of natural language processing (NLP) tasks.
Common benchmarks reveal significant variations in task types, dataset sizes, and domains. For instance, for fine-tuning datasets, training set sizes range from just 2.49k examples (RTE) to 105k examples (QNLI), while task types span from single-sentence classification (e.g., SST-2) to complex inference tasks (e.g., SWAG). 

This heterogeneity is further complicated by the different data types, ranging from movie reviews to news reports and Wikipedia articles.
Such diversity presents a unique challenge in developing pruning strategies that can effectively generalize across the spectrum of NLP tasks, underscoring the need for data pruning approaches in cross-dataset scenarios.

Existing cross-dataset pruning methods require computationally expensive sample ranking processes. These methods require training to be run on original data to collect pruning statistics, access to reference models and label information \cite{fayyaz2022bert,zayed2023deep}. 
As shown in Figure \ref{fig:fig1}, these methods typically take around 60 minutes to process standard datasets like SWAG, with more complex approaches or larger datasets demanding even more time. In contrast, our method achieves comparable or superior performance in mere seconds, regardless of dataset size or task complexity.
The superiority in time efficiency and performance of our method is shown in Figure \ref{fig:fig1}.

To tackle these problems, we introduce Swift Cross-Dataset Pruning (SCDP). Specifically, our approach introduces \textbf{Frequency Distance (FD)} score, in which we leverage TF-IDF embeddings combined with geometric median calculations to swiftly assess sample importance. This technique offers two significant advantages.
1) \textbf{Cross-Dataset Generalizability}: By using TF-IDF embeddings, our method captures the semantic importance of words across various NLU tasks and domains. The geometric median calculation then provides a task-agnostic measure of centrality in the embedding space. This combination ensures that our approach is universally adaptable across multiple datasets and task types, from language inference to reasoning and beyond.
2) \textbf{Computational Efficiency}: Unlike existing cross-dataset pruning methods that often require computationally expensive processes such as model training, access to reference models, or label information \cite{fayyaz2022bert, zayed2023deep}, our approach allows for rapid evaluation of sample importance. The TF-IDF and geometric median calculations can be performed efficiently on raw text data, significantly reducing the computational overhead typically associated with sample ranking processes.

Furthermore, we apply \textbf{dataset size-adaptive pruning} to ensure diversity for two distinct scenarios.
For smaller datasets, we retain samples far from the geometric median, preserving outliers and edge cases to maintain diversity by keeping "unusual" examples. 
For larger datasets, we select samples from each stratum to maintain a balanced representation of the data distribution while significantly reducing the dataset size.

Our main contributions are:
\begin{itemize}[noitemsep,topsep=0pt,parsep=0pt,partopsep=0pt]
    \item We propose  Frequency Distance, a score that uses TF-IDF embeddings and geometric median to swiftly rank samples.
     \item We propose dataset size-adaptive pruning to enhance adaptability in cross-dataset setting.
    \item We conduct extensive experiments on six diverse datasets, encompassing various tasks such as paraphrase identification, natural language inference, and reasoning. Our experiments span a range of dataset sizes, demonstrating the superior efficiency and performance of our method in cross-dataset settings.
\end{itemize}

\section{Related Work}

\textbf{Dataset Pruning for Vision Dataset.} Early works in dataset pruning focus its application in computer vision tasks. \citet{toneva2018empirical} define forgetting as the number of transitions from correct prediction to incorrect prediction of a sample in training, and use this to rank samples. \citet{paul2021deep} propose to use EL2N and GraNd scores obtained from training period to rank samples. AUM~\cite{pleiss2020identifying} is a metric that calculates the difference between the logit of the ground truth label and the highest other logit. \citet{coleman2020selection} propose to use a small proxy model to obtain presentation for dataset pruning. These works require training on original data and a pre-trained model to obtain sample ranking metrics and are computationally expensive for large-scale models. On sampling from ranking metrics, \citet{zhengcoverage} propose coverage-centric coreset selection, an algorithm based on stratified sampling for dataset pruning and achieve better performance at high pruning rates for image classification tasks. \citet{xia2022moderate} selects data points with scores that are close to the score median  to build a moderate coreset.

\textbf{Dataset Pruning for Language Model Pre-training.} Most previous works on data efficiency for language tasks focus on the pre-training phase of LMs. Common approaches for this task are language filtering \cite{wenzek2020ccnet,raffel2020exploring,xue2021mt5,laurenccon2022bigscience}, heuristic filtering \cite{rae2021scaling,xue2021mt5}, data quality assessment \cite{du2022glam,marion2023less}, data deduplication \cite{lee2022deduplicating,abbas2023semdedup,tirumala2024d4}, toxic or explicit content filtering \cite{jansen2022perplexed,subramani2023detecting,maharana2024mathbbd}. These methods address issues like irrelevant content and data redundancy before the model learns from the data, which is crucial in LM pre-training. However, it is difficult to apply these methods to cross-dataset scenarios, due to differences in task target, use-case, dataset pruning criterias. Dataset pruning in cross-dataset settings is more difficult due to the wide range of tasks, dataset sizes involved.

\textbf{Dataset Pruning for Language Model Fine-tuning.} For fine-tuning phase of pre-trained LMs, \citet{attendu2023nlu} and \citet{fayyaz2022bert} study the application of EL2N and GraNd scores to fine-tuning transformer-based language models. \citet{zayed2023deep} propose a new metric based on EL2N which also use model logits to prune samples for fairness. \citet{yang2024smalltolarge} use training trajectories from small models to select samples for dataset pruning. \citet{chenalpagasus} try to use strong LLMs such as ChatGPT to rate the quality of samples. \citet{maharana2024mathbbd} create a graph for the whole dataset, in which each node represents a sample and is initiliazed with difficulty score from model training. These methods require extensive training time on original data and depend on models to assess data importance, making them computationally expensive. 

\section{Methodology}

\subsection{Task Description and Formulation}
Given a training set of a fine-tuning task $S=\left(x_i, y_i\right)^N_{i=1}$, where $N \in \mathbb{N}$, $x_i$ represents the $i$-th input and $y_i$ is its true label. Pruning rate $r \in \mathbb{R}$, where $0 < r < 1$ represents the portion of dataset that will be removed. The objective of dataset pruning is to identify a coreset $S_c \subset S$ where $|S_c|=(1-r)|S|$, and when the model is trained on $S_c$, it may still retain the highest possible performance on the test set.

\subsection{Sample Vectorization via TF-IDF}
Term Frequency-Inverse Document Frequency (TF-IDF) \cite{sparck1972statistical} is a numerical measure that indicates the importance of a word within a specific document in relation to its occurrence across a collection of documents, or corpora. TF-IDF is ideal in this scenario because it captures the significance of terms relative to the entire dataset while being fast and scalable. Unlike transformer-based embeddings like BERT \cite{devlin-etal-2019-bert}, which focus on contextual similarity and are computationally intensive, TF-IDF emphasizes term frequency and rarity, making it efficient for identifying unique and informative samples. This allows for effective pruning strategies that preserve diverse and representative samples.

We use TF-IDF for unigrams to assess the significance of a term in a sample by considering how frequently it appears in that sample and how infrequently it appears across other samples in the dataset. From the dataset $S$, we obtain the vocabulary $\mathcal{V} = \{w_0, w_1, ..., w_n\}$ which contains all terms from all samples in dataset $S$, $w_j$ denotes the $j$-th term in the vocabulary.

The term frequency for term $w_j$ in sample $x_i$, denoted as $\mathbf{tf}_{i, j} \in \mathbb{R}$, is calculated as: 
\begin{equation}
    \mathbf{tf}_{i, j} = \frac{f(w_j, x_i)}{\sum_{w_k \in x_i}f(w_k, x_i)},
\end{equation}
where $f(w_k, x_i)$ is the number of occurrences of term $w_k$ in sample $x_i$.

By concatenating all the term frequencies of each term in sample $x_i$, we get the term frequency vector $\mathbf{tf}_i \in \mathbb{R}^n$ of sample $x_i$ as follows:
\begin{equation}
    \mathbf{tf}_i = \mathbf{tf}_{i, 0} \oplus \mathbf{tf}_{i, 1} \oplus ... \oplus \mathbf{tf}_{i, n}.
\end{equation}

The inverse document frequency of a term in the dataset $\mathbf{idf}_j \in \mathbb{R}$ is computed as: 
\begin{equation}
  \mathbf{idf}_j=\log \frac{N}{1+\text{df}\left(w_j\right)}.
\end{equation}

where df$(w_j)$ is the number of samples from the dataset that contains the term $w_j$.

By concatenating all the inverse document frequencies of each term, we get the term frequency vector $\mathbf{idf} \in \mathbb{R}^n$ as follows:
\begin{equation}
    \mathbf{idf} = \mathbf{idf}_{0} \oplus \mathbf{idf}_{1} \oplus ... \oplus \mathbf{idf}_{n},
\end{equation}

Each sample $x_i$ is represented as a vector of TF-IDF scores. The TF-IDF vector $\mathbf{t}_i \in \mathbb{R}^{n}$ for sample $x_i$ is given by:
\begin{equation}
    \mathbf{t}_i = \mathbf{tf}_i \odot \mathbf{idf},
\label{eq:TD-IDF}
\end{equation}
where $\odot$ denotes the element-wise product operation of two vectors.

\subsection{Frequency Distance Score}
Due to the sparse nature of TF-IDF representations in datasets, clustering methods, as previously used for dataset pruning \cite{das2023deft,yang2024smalltolarge}, cannot be applied. Therefore, we propose Frequency Distance (FD), a new distance-based scoring metric with geometric median, which calculates distance of each sample in the embedding space to the geometric median. This score represents the relative position of each sample with regard to the geometric median of the whole dataset in the embedding space.

From the set of $N$ points $\{\textbf{t}_0$, $\textbf{t}_1$,...,$\textbf{t}_N\}$ which represent the embeddings of each document, we find the geometric median point $\mathbf{g}^* \in \mathbb{R}^n$ that minimizes the sum of L2 distances to every point:
\begin{equation}
\begin{aligned}
    \mathbf{g}^* & = \underset{{\mathbf{g}\in \mathbb{R}^n}}{\arg\min}f(\mathbf{g}),\\
    \text{ where } f(\mathbf{g}) & = \sum_{i \in [1,N]} ||\mathbf{g}-\mathbf{t}_i||_2.
\end{aligned}
\end{equation}

\renewcommand{\algorithmicrequire}{\textbf{Inputs:}}
\renewcommand{\algorithmicensure}{\textbf{Outputs:}}

\begin{algorithm}[t]
\caption{Dataset-size adaptive pruning}\label{alg:ss}
\begin{algorithmic}
\Require $S = \{(x_i, y_i)\}_{i=1}^n$: original dataset; $D = \{\text{FD}(x_i)\}_{i=1}^n$: set of calculated Frequency Distance scores; $r$: dataset pruning rate; $k$: the number of strata.
\Ensure $S_c$: the selected coreset

\Function{SizeAdaptivePruning}{$S$, $D$, $r$, $k$}
\If{$(1-r)|S| > 1500$}
    \State $R_1, R_2,...,R_k \gets$ splits scores from $D$ into $k$ ranges with even width
    \State $\mathcal{B} \gets$ \{$\mathbb{B}_i$: $\mathbb{B}_i$ consists of samples whose scores are in $R_i$\}
    \State $m \gets n \times (1-r)$
    \State $S_c \gets \varnothing$
    \While{$\mathcal{B} \neq \varnothing$}
    
        $\mathcal{B}_{min} \gets \underset{\mathbb{B}\in\mathcal{B}}{\arg\min}|\mathbb{B}|$
        
        $m_B \gets \min\{|\mathbb{B}_{min}|, \lfloor \frac{m}{|\mathcal{B}|} \rfloor\}$
        
        $S_B \gets$ randomly sample $m_B$ samples from $\mathbb{B}_{min}$
        
        $S_c \gets S_c \cup S_B$
        
        $\mathcal{B} \gets \mathcal{B} \setminus \{\mathbb{B}_{min}\}$
        
        $m \gets m - m_B$
    \EndWhile
\ElsIf{$(1-r)|S| \leq 1500$}
    \State $D' \gets$ \texttt{argsort}$(D)$
    \State $S_c \gets D'[(1-r)|S|:]$
\EndIf
\EndFunction
\end{algorithmic}
\end{algorithm}

Computing the geometric median is challenging, and no linear time algorithm currently exists \cite{bajaj1988algebraic}. Consequently, we employ an approximation technique proposed by \citet{vardi2000multivariate} to estimate the geometric median. This approach yields an $\epsilon$-accurate geometric median, satisfying the condition $f(\mathbf{g}_\epsilon) \leq (1+\epsilon)f(\mathbf{g}^*)$.

For each sample $x_i$ in the dataset, we obtain FD score by calculating the L2 distance of its embedding to the geometric median $d_i \in \mathbb{R}$ as follows: 
\begin{equation}
    \text{FD}(x_i) = ||\mathbf{t}_i-\mathbf{g}_{\epsilon}||_2.
\label{eq:fd}
\end{equation}

For each sample, we use Eq.~\ref{eq:fd} to calculate its score, to obtain the set of score of every sample $D = \{\text{FD}(x_i)\}_{i=1}^N$. This metric is used to evaluate data points in the training set to perform dataset pruning.

\begin{table}[!t]
\centering
\small
\begin{tabularx}{\columnwidth}{*{4}{>{\centering\arraybackslash}X}}
\toprule
\textbf{Dataset} & \textbf{Metric} & \textbf{Task} & \textbf{Size} \\ 
\midrule
RTE & Accuracy & NLI & 2.49k \\
\midrule
MRPC & Accuracy & Paraphrase & 3.67k \\
\midrule
CoLA & Matthews corr. & Grammatical Acceptability & 8.55k\\
\midrule
SST-2 & Accuracy & Sentiment Analysis & 67.3k \\
\midrule
SWAG & Accuracy & Reasoning & 73.5k \\
\midrule
QNLI & Accuracy & QA/NLI & 105k \\
\bottomrule
\end{tabularx}
\caption{Evaluation metric, task and original size of train set of the datasets used in experimental evaluation. QA: Question Answering, NLI: Natural Language Inference.}
\label{tab:datasets}
\end{table}

\subsection{Dataset size-adaptive pruning}

In our cross-dataset setting, to ensure performance across diverse scales of datasets, we present dataset size-adaptive pruning, a novel sampling method to choose samples from set of FD scores, as presented in Algorithm \ref{alg:ss}. By applying dataset size-adaptive pruning, we ensure diversity for two distinct scenarios.

For smaller datasets, every sample could potentially carry unique information crucial for model performance. We retain samples far from the geometric median, preserving outliers and edge cases. This maintains diversity by keeping "unusual" examples that, while rare, are crucial for comprehensive model training and incentivize the understanding of rare or complex patterns.

For larger datasets, the challenge lies in maintaining a balanced representation of the data distribution while significantly reducing the dataset size. By selecting samples from each stratum following \citet{zhengcoverage}, we ensure a diverse range of examples in the pruned dataset, from typical central cases to unique peripheral ones. Dataset size-adaptive pruning is applied as follows:
\begin{itemize}[noitemsep,topsep=0pt,parsep=0pt,partopsep=0pt,leftmargin=*]
    \item Case 1: For \textbf{small post-pruning coreset size}, where $(1-r)|S| \leq 1500$, keeping furthest samples is the preferred strategy. From the set of calculated FD scores $D$, this strategy keeps the samples furthest to geometric median as coreset $S_c$.
    \item Case 2: For \textbf{middle to large post-pruning coreset size}, where $(1-r)|S| > 1500$, since we have enough representative samples in each stratum, we use stratified sampling to ensure that the pruned dataset retains a representative mix of samples across different strata.
\end{itemize}

\section{Experiments}
\subsection{Experiment Settings}

\begin{table}[t]
    \small
    \centering
    \begin{tabularx}{\columnwidth}{
        >{\centering\arraybackslash}X
        >{\centering\arraybackslash}X
        >{\centering\arraybackslash}X
        >{\centering\arraybackslash}p{1.1cm}
        >{\centering\arraybackslash}X
        >{\centering\arraybackslash}X
    }
    \toprule
    PR & EL2N & AUM & Forgetting & CCS & Ours \\
    \midrule
    \multicolumn{6}{c}{RTE} \\
    \midrule
    50\% & 45.84 & 45.72 & 47.53 & 50.78 & \textbf{55.83} \\
    70\% & 43.96 & 45.00 & 45.12 & 48.49 & \textbf{57.40} \\
    \midrule
    \multicolumn{6}{c}{MRPC} \\
    \midrule
    50\% & 68.54 & 68.38 & 74.58 & 77.77 & \textbf{83.00} \\
    70\% & 68.38 & 68.38 & 70.09 & 71.64 & \textbf{75.73} \\
    \midrule
    \multicolumn{6}{c}{CoLA} \\
    \midrule
    50\% & 12.90 & 0.00 & 43.94 & \textbf{46.38} & 45.30 \\
    70\% & 0.01 & 0.00 & 4.05 & 36.86 & \textbf{43.39} \\
    \midrule
    \multicolumn{6}{c}{SST-2} \\
    \midrule
    50\% & 90.51 & \textbf{90.97} & 90.75 & 90.02 & 90.74 \\
    70\% & 88.79 & 88.99 & 89.90 & 89.52 & \textbf{90.21} \\
    \midrule
    \multicolumn{6}{c}{SWAG} \\
    \midrule
    50\% & 55.76 & 50.54 & 61.40 & 64.57 & \textbf{65.43} \\
    70\% & 28.14 & 23.83 & 53.17 & 61.82 & \textbf{63.54} \\
    \midrule
    \multicolumn{6}{c}{QNLI} \\
    \midrule
    50\% & 83.88 & 84.47 & 86.50 & 86.50 & \textbf{87.19} \\
    70\% & 66.53 & 42.68 & 75.94 & 83.88 & \textbf{85.52} \\
    \bottomrule
    \end{tabularx}
    \caption{Overall result of baselines and our method. Pruning rate (PR) is the percentage of data that is removed from full training data during dataset pruning. The best results are highlighted in \textbf{bold}.}
    \label{tab:results}
\end{table}

\textbf{Evaluation Datasets.} We experimented on six natural language understanding datasets, including RTE, QNLI, CoLA, MRPC, SST-2 from the GLUE benchmark \cite{wang2019glue}, SWAG \cite{zellers2018swag}. The task and evaluation metric of each dataset is listed in Table \ref{tab:datasets}. The selected datasets have a large variance in size of samples and has a wide variety of tasks, which demonstrates the universal effectiveness of our method. 

\textbf{Model settings.} We fine-tune pre-trained DistilBERT~\cite{sanh2019distilbert} in all experiments. A task-specific head is added to the final layer of DistilBERT. First, we use the proposed method to prune and obtain the remaining coreset and fine-tune DistilBERT with the coreset. We fine-tune DistilBERT on $3$ epochs with an initial learning rate of $5e-5$ with cosine annealing scheduler \cite{loshchilov2016sgdr}, batch size $32$, using AdamW optimizer \cite{loshchilovdecoupled}.

\textbf{Pruning settings.} For geometric median approximation, we use $\epsilon=1e-5$. For stratified sampling, we set the number of strata to $k=100$.

\begin{table}[t]
    \centering
    \small
    \begin{tabularx}{\linewidth}{l*{4}{>{\centering\arraybackslash}X}}
    \toprule
    Pruning rate & 10\% & 30\% & 50\% & 70\% \\
    \midrule
    \multicolumn{5}{c}{RTE} \\
    \midrule
    Sentence-BERT & 57.03 & \textbf{57.88} & \textbf{55.83} & 49.81 \\
    \rowcolor{gray!25} TF-IDF & \textbf{61.85} & 57.39 & \textbf{55.83} & \textbf{57.40} \\
    \midrule
    \multicolumn{5}{c}{MRPC} \\
    \midrule
    Sentence-BERT & 84.47 & 83.33 & 81.78 & 72.87  \\
    \rowcolor{gray!25} TF-IDF & \textbf{84.55} & \textbf{83.41} & \textbf{83.00} & \textbf{75.73} \\
    \midrule
    \multicolumn{5}{c}{CoLA} \\
    \midrule
    Sentence-BERT & \textbf{49.53} & 47.39 & \textbf{45.65} & 41.58 \\
    \rowcolor{gray!25} TF-IDF & 49.43 & \textbf{48.09} & 45.30 & \textbf{43.39} \\
    \midrule
    \multicolumn{5}{c}{SST-2} \\
    \midrule
    Sentence-BERT & 90.63 & 90.71 & 90.21 & 89.44 \\
    \rowcolor{gray!25} TF-IDF & \textbf{90.97} & \textbf{91.13} & \textbf{90.74} & \textbf{90.21} \\
    \midrule
    \multicolumn{5}{c}{SWAG} \\
    \midrule
    Sentence-BERT & 67.09 & 66.63 & 64.57 & 63.47 \\
    \rowcolor{gray!25} TF-IDF & \textbf{67.45} & \textbf{66.68} & \textbf{65.43} & \textbf{63.54} \\
    \midrule
    \multicolumn{5}{c}{QNLI} \\
    \midrule
    Sentence-BERT & 89.01 & 87.80 & \textbf{87.21} & 85.30 \\
    \rowcolor{gray!25} TF-IDF & \textbf{89.10} & \textbf{88.27} & 87.19 & \textbf{85.52} \\
    \bottomrule
    \end{tabularx}
    \caption{Ablation study for embedding methods. The best results are highlighted in \textbf{bold}.}
    \label{tab:ablation_embedding}
\end{table}

\textbf{Baselines.} We compare our method against five baselines, four of which are SOTA methods for dataset pruning. We use \textbf{(1) Random}: we randomly select samples to form the coreset, \textbf{(2) AUM} \cite{pleiss2020identifying}, \textbf{(3) EL2N} \cite{paul2021deep}, \textbf{(4) Forgetting} \cite{toneva2018empirical}, \textbf{(5) CCS} \cite{zhengcoverage}: coverage-centric coreset selection with AUM as the importance score. All experiments are run three times and average score is reported in this paper.

\subsection{Experimental Results}

\textbf{Main Experiments.} The evaluation results of the baseline methods are compared to our proposed method. We conduct experiments on multiple pruning rates to investigate how our method perform at different data compression rates. In Table \ref{tab:results}, we present the performance on all datasets at 50\% and 70\% pruning rates. Full results at 10\%, 30\%, 50\% and 70\% over all datasets are presented in Appendix \ref{full_results_apx}. In overall, our proposed method performs the best compared to all other baselines methods and outperforms baseline methods.

When compared to SOTA baselines, our method has best performance overall. Our method consistently outperforms state-of-the-art baselines AUM, EL2N, Forgetting and CCS. For high post-pruning coreset size, distance-based stratified sampling is particularly effective because it ensures that the pruned dataset remains both diverse and informative, eliminating redundant data while preserving the core structure of the dataset. At small post-pruning coreset size, furthest samples prove to be effective, since most distant from the central tendency can help maintain the diversity and richness of the data.

\begin{table}[!t]
    \centering
    \small
    \begin{tabularx}{\linewidth}{l*{4}{>{\centering\arraybackslash}X}}
    \toprule
    Pruning rate & 10\% & 30\% & 50\% & 70\% \\
    \midrule
    \multicolumn{5}{c}{RTE} \\
    \midrule
    Random & 58.06 & 54.75 & 53.90 & 54.75 \\
    \rowcolor{gray!25} Our method & \textbf{61.85} & \textbf{57.39} & \textbf{55.83} & \textbf{57.40} \\
    \midrule
    \multicolumn{5}{c}{SWAG} \\
    \midrule
    Random & 65.20 & 64.75 & 63.96 & 62.88 \\
    \rowcolor{gray!25} Our method & \textbf{67.45} & \textbf{66.68} & \textbf{65.43} & \textbf{63.54} \\
    \midrule
    \multicolumn{5}{c}{QNLI} \\
    \midrule
    Random & 87.44 & 87.32 & 86.06 & 85.27 \\
    \rowcolor{gray!25} Our method & \textbf{89.10} & \textbf{88.27} & \textbf{87.19} & \textbf{85.52} \\
    \bottomrule
    \end{tabularx}
    \caption{Comparison with random selection. The best results are highlighted in \textbf{bold}.}
    \label{tab:results_random}
\end{table}

\begin{table}[t]
    \centering
    \small
    \begin{tabularx}{\linewidth}{l*{5}{>{\centering\arraybackslash}X}}
    \toprule
    Model & BERT & ALBERT & XLNet & RoBERTa \\
    \midrule
    Random & \underline{57.03} & 52.70 & \textbf{61.01} & \underline{59.92} \\
    EL2N & 39.71 & 45.84 & 47.65 & 44.40 \\
    AUM & 43.32 & 48.01 & 50.54 & 52.70 \\
    Forgetting & 44.76 & \underline{55.59} & 50.54 & 52.70 \\
    CCS & 56.67 & 52.70 & 53.79 & 50.54 \\
    \rowcolor{gray!25} Ours & \textbf{61.37} & \textbf{57.40} & \underline{59.20} & \textbf{56.67} \\
    \bottomrule
    \end{tabularx}
    \caption{Experiments on other models on RTE dataset at 70\% pruning rate.  The best and the second best results are highlighted in \textbf{bold} and \underline{underlined}, respectively.}
    \label{tab:other_models}
\end{table}

\begin{figure*}[!ht]
    \centering
    \small
    \includegraphics[width=0.9\linewidth]{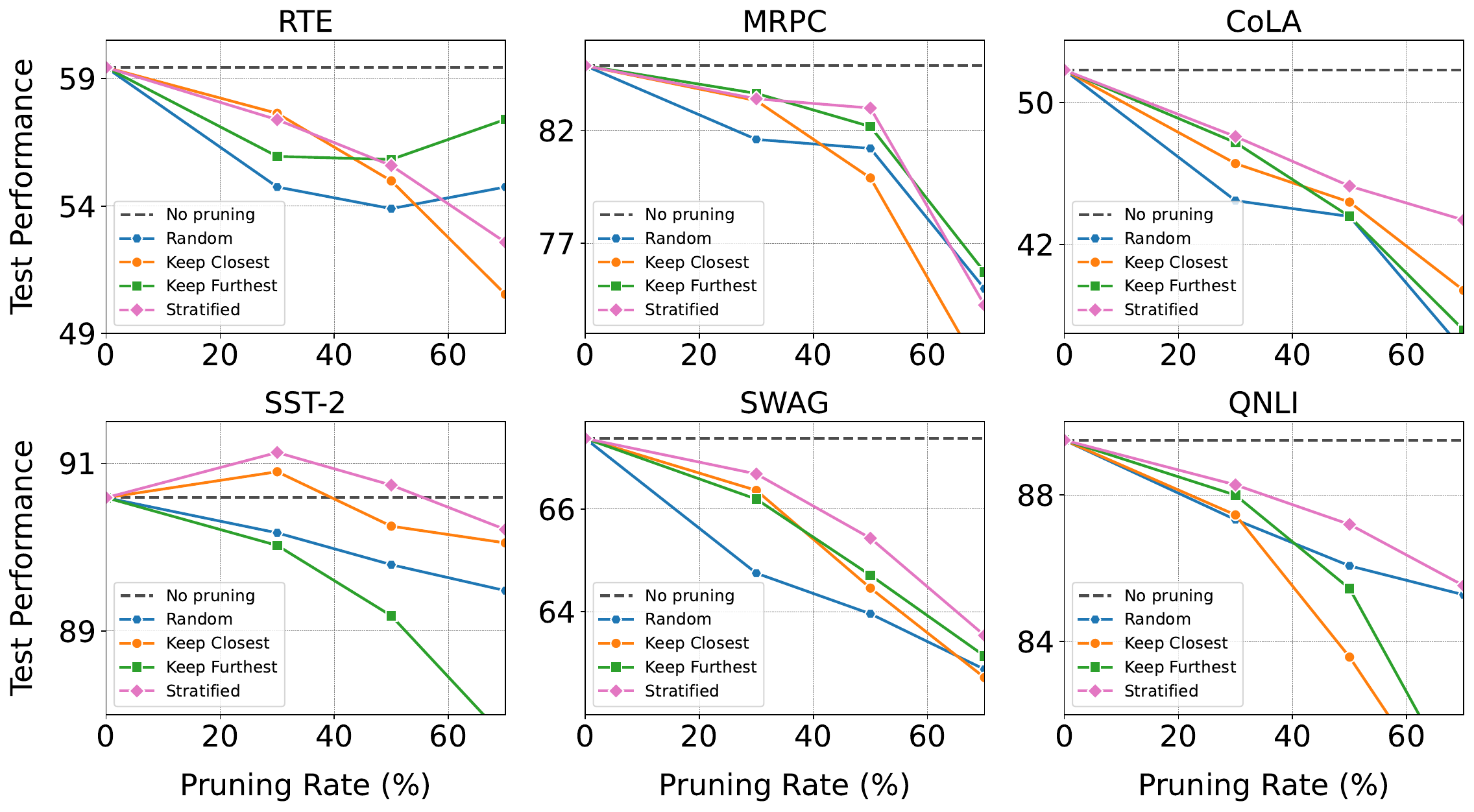}
    \caption{Results of pruning strategies from set of distance-based scores.}
    \label{fig:ablation_sampling}
\end{figure*}

Comparison between our method and random baseline is shown in Table \ref{tab:results_random}. At 10\% pruning rate, the overall improvement to random is 2.57\%, and at 70\% pruning rate, the overall improvement to random is 1.19\%.

\begin{figure}[!t]
    \centering
    \includegraphics[width=\linewidth]{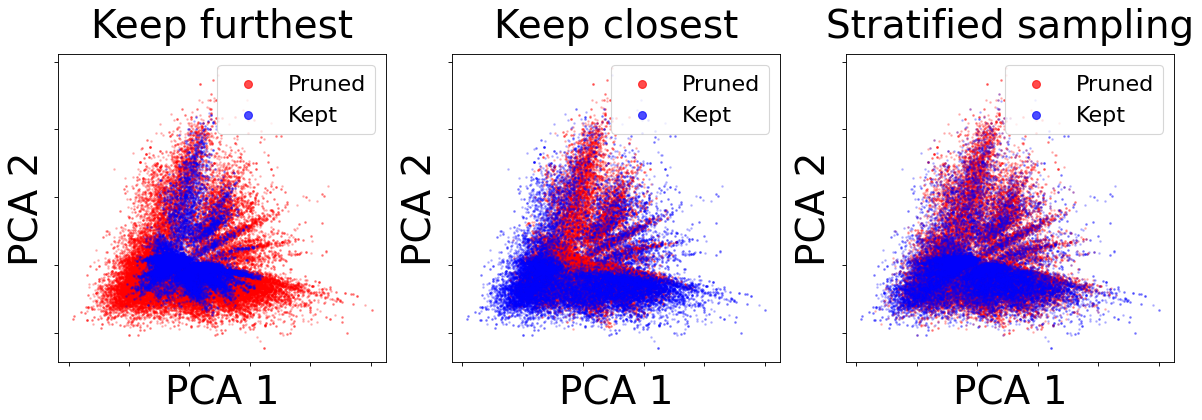}
    \caption{PCA Plot of selected data points with regard to full training set for SWAG dataset at 70\% pruning rates for different pruning strategies.}
    \label{fig:pca_swag}
\end{figure}

We tested our methods on other language models, specifically BERT \cite{devlin-etal-2019-bert}, ALBERT \cite{Lan2020ALBERT:}, XLNet \cite{DBLP:conf/nips/YangDYCSL19} and RoBERTa \cite{liu2019roberta} on the RTE dataset to evaluate the robustness of our methods, as shown in Table \ref{tab:other_models}. Our method consistently perform well across all models, which prove that our method is not a phenomenon with respect to a particular language model.

\textbf{Dataset size-adaptive pruning.}
In Figure \ref{fig:ablation_sampling}, we compare three different pruning strategies from distance-based scores: only keeping the closest samples to geometric median, only keeping the furthest samples to geometric median, and using stratified sampling to select the coreset.

With regard to the proposed dataset size-adaptive pruning, there are three cases where keeping furthest samples are applied: for MRPC dataset at 70\% pruning rate, and for RTE dataset at 50\% and 70\% pruning rates. For these scenarios, furthest samples prove to be more effective coresets, producing higher performance than stratified sampling. For coresets with bigger size, stratified sampling prove to be the best strategy compared to the other methods, since it maintains the distribution characteristics of the original dataset.

Keeping closest samples method still retain good performance at 30\% pruning rate. However, at higher pruning rates, its performance drop quickly, for most datasets it may perform worse than random pruning.

\subsection{Further Analysis}

\begin{figure}[!t]
    \centering
    \includegraphics[width=\linewidth]{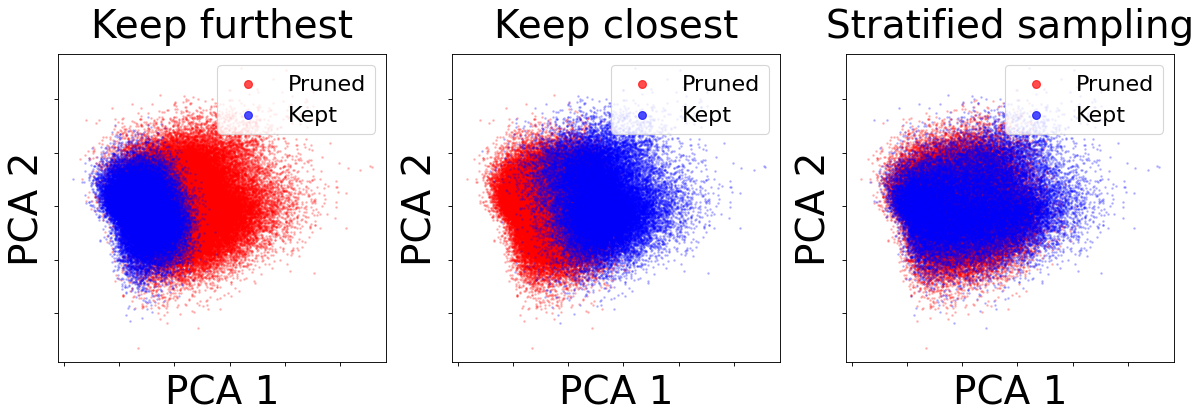}
    \caption{PCA Plot of selected data points with regard to full training set for QNLI dataset at 70\% pruning rates for different pruning strategies.}
    \label{fig:pca_qnli}
\end{figure}

\textbf{TF-IDF embedding performs superior to Sentence-BERT embedding.} To validate the effectiveness of TF-IDF embedding, we performed experiments with Sentence-BERT \cite{sentencebert} embedding of samples. The performance of TF-IDF is superior compared to Sentence-BERT across all datasets, as shown in Table \ref{tab:ablation_embedding}. TF-IDF is superior to Sentence-BERT embeddings in this context because it directly captures the importance of individual words within each sample, emphasizing terms that are significant to the dataset. This makes TF-IDF particularly effective for identifying central and representative samples in the dataset, as it reflects the specific vocabulary and term frequency patterns of the data. In contrast, Sentence-BERT embeddings focus on capturing the overall semantic meaning of sentences, which obscure the importance of individual words and lead to less precise centrality measures.

\textbf{Stratified sampling maintains good coverage in embedding space.} Figure \ref{fig:pca_swag} and Figure \ref{fig:pca_qnli} demonstrate the selected data points and the pruned data points for 70\% pruning rate in 2 dimension space using principle component analysis \cite{wold1987principal}. Stratified sampling method works very well for medium to large size datasets since it is able to keep a balanced representation of the data distribution. Compared to keep furthest strategy or keep closest strategy, stratified sampling coreset covers much wider regions of the plot. Interestingly, in the keep furthest samples strategy, selected samples are placed relatively close to the center of the 2 dimensional plot in the SWAG dataset. This is explained by the sparsity of TF-IDF vectors of the furthest samples, where rare words are included and their lengths are often shorter.

\begin{CJK*}{UTF8}{gbsn}
\begin{table}[!t]
    \scriptsize
    \centering
    \begin{tabularx}{\linewidth}{>{\centering\arraybackslash}m{0.5cm} >{\raggedright\arraybackslash}m{\dimexpr\linewidth-3.9cm\relax} >
    {\centering\arraybackslash}m{0.9cm} >{\centering\arraybackslash}m{0.9cm}}
    \toprule
    ID & \multicolumn{1}{c}{Example} & FD & QuRating \\
    \midrule
    2015 & What does 基督徒 (pinyin: jīdū tú) mean? (pinyin: jīdū tú), literally "Christ follower." & 1.009 (99.99\%) & -1.22 (6.72\%) \\
    \midrule
    22851 & Which vowel remains distinct? ; /i/ remains distinct. & 1.008 (99.99\%) & -0.83 (11.90\%) \\
    \midrule
    10539 & Name one of the individuals considered the founding fathers of modern Cypriot art. Arguably the two founding fathers of modern Cypriot art were Adamantios Diamantis (1900--1994) who studied at London's Royal College of Art and Christopheros Savva (1924--1968) who also studied in London, at Saint Martin's School of Art. & 0.990 (53.77\%) & 3.92 (98.58\%) \\
    \midrule
    40675 & Where was the University of Paris located? The Left Bank was the site of the University of Paris, a corporation of students and teachers formed in the mid-12th century to train scholars first in theology, and later in canon law, medicine and the arts. & 0.966 (0.31\%) & 3.10 (95.28\%) \\
    \bottomrule
    \end{tabularx}
    \caption{Comparison to QuRating. The percentage shows a sample's percentile rank in the dataset.}
    \label{tab:example_qr}
\end{table}
\end{CJK*}

\textbf{Comparison of our Frequency Distance score with previous quality-based score.}
To compare our method with quality-based methods in pre-training dataset pruning, we use QuRating \cite{wettigqurating}, a method that utilizes a fine-tuned Sheared-Llama-1.3B model \cite{xiasheared} to judge the quality of text, and perplexity (PPL) of GPT-2 \cite{radford2019language}. Quality assessment is often used in pre-training dataset pruning to select data that resembles a high-quality corpus \cite{albalak2024a, wettigqurating,chowdhery2023palm,xie2023data}. For QuRating, high scoring samples are preferred, while for perplexity, low scoring samples are preferred. However, this approach is not applicable to cross-dataset pruning, since it hurts diversity of the coreset and harms model performance in this scenario. 

In Table \ref{tab:example_qr} and Table \ref{tab:example_ppl}, we show examples that are chosen by our stratified sampling strategy at 70\% pruning rate in the QNLI dataset and compare FD score to QuRating and perplexity scores. Our method selects samples ranging from low quality to high quality, ensuring a more diverse set of samples. In contrast, quality-based methods like QuRating or perplexity-based quality filtering prioritizes high quality samples, which can lead to a loss of important linguistic and contextual variety.

\section{Conclusion}

\begin{CJK*}{UTF8}{gbsn}
\begin{table}[!t]
    \scriptsize
    \centering
    \begin{tabularx}{\linewidth}{>{\centering\arraybackslash}m{0.5cm} >{\raggedright\arraybackslash}m{\dimexpr\linewidth-3.9cm\relax} >
    {\centering\arraybackslash}m{0.9cm} >{\centering\arraybackslash}m{0.9cm}}
    \toprule
    ID & \multicolumn{1}{c}{Example} & FD & PPL \\
    \midrule
    2015 & What does 基督徒 (pinyin: jīdū tú) mean? (pinyin: jīdū tú), literally "Christ follower." & 1.009 (99.99\%) & 34.19 (30.37\%) \\
    \midrule
    22851 & Which vowel remains distinct? ; /i/ remains distinct. & 1.008 (99.99\%) & 399.56 (99.81\%) \\
    \midrule
    10539 & Name one of the individuals considered the founding fathers of modern Cypriot art. Arguably the two founding fathers of modern Cypriot art were Adamantios Diamantis (1900--1994) who studied at London's Royal College of Art and Christopheros Savva (1924--1968) who also studied in London, at Saint Martin's School of Art. & 0.990 (53.77\%) & 31.06 (24.77\%) \\
    \midrule
    40675 & Where was the University of Paris located? The Left Bank was the site of the University of Paris, a corporation of students and teachers formed in the mid-12th century to train scholars first in theology, and later in canon law, medicine and the arts. & 0.966 (0.31\%) & 26.04 (16.06\%) \\
    \bottomrule
    \end{tabularx}
    \caption{Comparison to perplexity (PPL) of GPT-2.}
    \label{tab:example_ppl}
\end{table}
\end{CJK*}

In this paper, we introduce SCDP to enhance fine-tuning efficiency for NLP tasks in cross-dataset scenarios. We propose Frequency Distance, a sample ranking score that is computationally efficient and bypasses the need for expensive reference models or training on the full original data. Furthermore, we propose dataset size-adaptive pruning to improve adaptability across a diverse range of tasks and dataset sizes. Our extensive experiments across six datasets validate the effectiveness of this approach, demonstrating that our method achieves competitive performance while reducing computational costs. This work represents a significant step towards data-efficient training in NLP, particularly for fine-tuning in cross-dataset settings.

\section{Limitations}

While our method shows promising results, it has several limitations.
1) Task diversity: Although tested on six NLU tasks, the method's performance on more complex tasks such as text generation is unknown.
2) Theoretical grounding: While empirically effective, our work lacks a rigorous theoretical analysis explaining the superiority of TF-IDF with the geometric median for cross-dataset pruning.

\bibliography{custom}

\clearpage 

\appendix

\section{Experiments}

\label{sec:appendix_experiments}

\subsection{Dataset Descriptions}

We provide a detailed description of datasets used in our experiments below:

\begin{itemize}
    \item \textbf{RTE}. In the input, two text fragments are given. The task is to recognize whether the meaning of one text is entailed from the other text.
    \item \textbf{MRPC}. This dataset consists of sentence pairs automatically extracted from online news sources, with human annotations for whether the sentences in the pair are semantically equivalent.
    \item \textbf{CoLA}. This dataset consists of English acceptability judgments drawn from books and journal articles on linguistic theory. Each example is a sequence of words annotated with whether it is a grammatical English sentence.
    \item \textbf{SST-2}. This dataset consists of sentences from movie reviews and human annotations of their sentiment. The task is to predict the sentiment of a given sentence.
    \item \textbf{SWAG}. Given a partial text description, the model has to reason about the situation and anticipate what comes next by choosing one of multiple-choice text options.
    \item \textbf{QNLI}. This dataset is a question-answering dataset consisting of question-paragraph pairs, where one of the sentences in the paragraph contains the answer to the corresponding question. The task is to determine whether the context sentence contains the answer to the question.
\end{itemize}

\subsection{Data Preprocessing}
For each dataset, we preprocess the input before using TF-IDF embedding as follows:
\begin{itemize}
    \item \textbf{SWAG.} Concatenate sentence 1 and sentence 2 as input to get TF-IDF embedding, and not include 4 answer options.
    \item \textbf{SST-2.} The input sentence is the input to get TF-IDF embedding.
    \item \textbf{QNLI.} Concatenate question and answer as input to get TF-IDF embedding.
    \item \textbf{RTE.} Concatenate sentence 1 and sentence 2 as input to get TF-IDF embedding.
    \item \textbf{CoLA.} The input sentence is the input to get TF-IDF embedding.
    \item \textbf{MRPC.} The input sentence is the input to get TF-IDF embedding.
\end{itemize}

\subsection{Baseline Experiment Settings}
For \textbf{AUM}, \textbf{EL2N}, \textbf{CCS} baseline, we get training statistics in every epoch. That is, for each epoch, predictions on the training set will be used to calculate sample importance. Since we fine-tune DistilBERT on 3 epochs for every dataset, we obtain scores 3 times and use the mean as the final sample importance score. For \textbf{Forgetting} baseline, since 3 times are not enough to evaluate forgetting scores, we evaluate the model on the training set after iterations instead of epochs. That is, we evaluate the model on the training set after every fixed number of iterations to get the forgetting information of samples.

\subsection{Detailed Results}
\label{full_results_apx}

Table \ref{tab:detailed_result} describes the detailed results over 10\%, 30\%, 50\%, 70\% pruning rates on all datasets. The result of fine-tuning on the full dataset is shown in the rows with 0\% pruning rate

\begin{table*}[ht!]
    \centering
    \scriptsize
    \renewcommand{\arraystretch}{1.2}
    \begin{tabular}{>{\centering\arraybackslash}m{1.5cm}>{\centering\arraybackslash}m{1.9cm}>{\centering\arraybackslash}m{1.9cm}>{\centering\arraybackslash}m{1.9cm}>{\centering\arraybackslash}m{1.9cm}>{\centering\arraybackslash}m{1.9cm}>{\centering\arraybackslash}m{1.9cm}}
    \hline
    Pruning rate & \multicolumn{1}{c}{Random} & \multicolumn{1}{c}{EL2N} & \multicolumn{1}{c}{AUM} & \multicolumn{1}{c}{Forgetting} & \multicolumn{1}{c}{CCS} & \multicolumn{1}{c}{Ours} \\
    \hline
    &\multicolumn{6}{c}{RTE} \\
    \hline
    0\% & \multicolumn{6}{c}{$59.44_{\pm4.37}$} \\
    10\% & $58.06_{\pm1.50}$ & $53.67_{\pm0.91}$ & $54.63_{\pm3.34}$ & $55.95_{\pm2.73}$ & $58.12_{\pm1.91}$ & $\mathbf{61.85_{\pm1.82}}$ \\
    30\% & $54.75_{\pm0.91}$ & $48.49_{\pm3.24}$ & $46.08_{\pm1.37}$ & $48.13_{\pm4.93}$ & $55.71_{\pm2.92}$ & $\mathbf{57.39_{\pm0.62}}$ \\
    50\% & $53.90_{\pm2.46}$ & $45.84_{\pm0.96}$ & $45.72_{\pm0.76}$ & $47.53_{\pm0.21}$ & $50.78_{\pm2.53}$ & $\mathbf{55.83_{\pm0.62}}$ \\
    70\% & $54.75_{\pm0.55}$ & $43.96_{\pm0.50}$ & $45.00_{\pm1.37}$ & $45.12_{\pm1.65}$ & $48.49_{\pm0.55}$ & $\mathbf{57.40_{\pm2.35}}$ \\
    \hline
    &\multicolumn{6}{c}{MRPC} \\
    \hline
    0\% & \multicolumn{6}{c}{$84.88_{\pm0.14}$} \\
    10\% & $82.43_{\pm0.71}$ & $84.15_{\pm1.02}$ & $\mathbf{85.53_{\pm0.65}}$ & $84.80_{\pm0.74}$ & $83.98_{\pm0.99}$ & $84.55_{\pm1.91}$ \\
    30\% & $81.61_{\pm1.77}$ & $80.06_{\pm4.41}$ & $78.84_{\pm4.50}$ & $82.84_{\pm2.01}$ & $\mathbf{84.31_{\pm1.37}}$ & $83.41_{\pm0.14}$ \\
    50\% & $81.21_{\pm1.58}$ & $68.54_{\pm0.28}$ & $68.38_{\pm0.00}$ & $74.58_{\pm3.20}$ & $77.77_{\pm3.33}$ & $\mathbf{83.00_{\pm0.57}}$ \\
    70\% & $74.99_{\pm4.43}$ & $68.38_{\pm0.00}$ & $68.38_{\pm0.00}$ & $70.09_{\pm0.88}$ & $71.64_{\pm1.63}$ & $\mathbf{75.73_{\pm3.78}}$ \\
    \hline
    &\multicolumn{6}{c}{CoLA} \\
    \hline
    0\% & \multicolumn{6}{c}{$51.83_{\pm1.73}$} \\
    10\% & $47.31_{\pm0.51}$ & $\mathbf{51.46_{\pm2.38}}$ & $50.50_{\pm0.93}$ & $49.72_{\pm2.82}$ & $51.43_{\pm2.54}$ & $49.43_{\pm1.09}$ \\
    30\% & $44.48_{\pm1.08}$ & $45.54_{\pm4.07}$ & $44.51_{\pm3.34}$ & $48.57_{\pm2.02}$ & $47.95_{\pm0.89}$ & $\mathbf{48.09_{\pm0.98}}$ \\
    50\% & $43.59_{\pm2.14}$ & $12.90_{\pm12.94}$ & $0.00_{\pm0.00}$ & $43.94_{\pm0.43}$ & $\mathbf{46.38_{\pm0.43}}$ & $45.30_{\pm0.43}$ \\
    70\% & $36.05_{\pm1.33}$ & $0.01_{\pm0.01}$ & $0.00_{\pm0.00}$ & $4.05_{\pm0.43}$ & $36.86_{\pm0.43}$ & $\mathbf{43.39_{\pm0.43}}$ \\
    \hline
    &\multicolumn{6}{c}{SST-2} \\
    \hline
    0\% & \multicolumn{6}{c}{$90.59_{\pm0.50}$} \\
    10\% & $90.32_{\pm0.78}$ & $90.71_{\pm0.20}$ & $90.73_{\pm0.32}$ & $90.44_{\pm0.27}$ & $\mathbf{91.05_{\pm0.11}}$ & $90.97_{\pm0.48}$ \\
    30\% & $90.17_{\pm0.54}$ & $\mathbf{91.43_{\pm0.24}}$ & $91.20_{\pm0.33}$ & $90.74_{\pm0.48}$ & $90.97_{\pm0.78}$ & $91.13_{\pm0.58}$ \\
    50\% & $89.79_{\pm0.80}$ & $90.51_{\pm0.84}$ & $\mathbf{90.97_{\pm0.35}}$ & $90.75_{\pm0.18}$ & $90.02_{\pm0.53}$ & $90.74_{\pm0.27}$ \\
    70\% & $89.48_{\pm0.17}$ & $88.79_{\pm1.09}$ & $88.99_{\pm0.64}$ & $89.90_{\pm0.64}$ & $89.52_{\pm0.47}$ & $\mathbf{90.21_{\pm0.29}}$ \\
    \hline
    &\multicolumn{6}{c}{SWAG} \\
    \hline
    0\% & \multicolumn{6}{c}{$67.37_{\pm0.23}$} \\
    10\% & $65.20_{\pm0.16}$ & $67.20_{\pm0.18}$ & $66.85_{\pm1.06}$ & $67.22_{\pm0.32}$ & $67.14_{\pm0.09}$ & $\mathbf{67.45_{\pm0.32}}$ \\
    30\% & $64.75_{\pm0.22}$ & $65.16_{\pm0.09}$ & $65.80_{\pm0.70}$ & $66.18_{\pm0.10}$ & $66.32_{\pm0.51}$ & $\mathbf{66.68_{\pm0.41}}$ \\
    50\% & $63.96_{\pm0.16}$ & $55.76_{\pm1.68}$ & $50.54_{\pm2.58}$ & $61.40_{\pm0.41}$ & $64.57_{\pm0.16}$ & $\mathbf{65.43_{\pm0.39}}$ \\
    70\% & $62.88_{\pm0.35}$ & $28.14_{\pm2.20}$ & $23.83_{\pm0.80}$ & $53.17_{\pm1.13}$ & $61.82_{\pm0.21}$ & $\mathbf{63.54_{\pm0.38}}$ \\
    \hline
    &\multicolumn{6}{c}{QNLI} \\
    \hline
    0\% & \multicolumn{6}{c}{$89.49_{\pm0.38}$} \\
    10\% & $87.44_{\pm0.38}$ & $89.10_{\pm0.35}$ & $89.41_{\pm0.31}$ & $\mathbf{89.19_{\pm0.18}}$ & $88.99_{\pm0.73}$ & $89.10_{\pm0.06}$ \\
    30\% & $87.32_{\pm0.59}$ & $87.38_{\pm0.88}$ & $88.61_{\pm0.07}$ & $\mathbf{88.80_{\pm0.53}}$ & $87.95_{\pm0.33}$ & $88.27_{\pm0.34}$ \\
    50\% & $86.06_{\pm0.42}$ & $83.88_{\pm0.69}$ & $84.47_{\pm0.86}$ & $86.50_{\pm1.31}$ & $86.50_{\pm0.70}$ & $\mathbf{87.19_{\pm0.07}}$ \\
    70\% & $85.27_{\pm0.43}$ & $66.53_{\pm0.43}$ & $42.68_{\pm0.21}$ & $75.94_{\pm1.96}$ & $83.88_{\pm0.68}$ & $\mathbf{85.52_{\pm1.02}}$ \\
    \hline

    \end{tabular}
    \caption{Overall result with standard deviation.}
    \label{tab:detailed_result}
\end{table*}

\section{Examples}

\subsection{Quality-based Comparison}

In Table 9-13, we list examples chosen by our method at 70\% pruning rate and compute their quality-based score with QuRating and Perplexity of GPT-2. Examples show that our method is able to keep the diversity of samples based on quality and perplexity, which benefits the model to generalize to linguistic and contextual features.

\subsection{Examples of Furthest Samples and Closest Samples to Geometric Median}

Text input of the samples furthest to the geometric median and samples closest to the geometric median of the QNLI dataset are displayed in Table \ref{tab:example}. By applying FD score, we can obtain the most distinctive samples by choosing samples with highest FD score. On the other hand, samples nearest to geometric median tend to have similar terms and topics. Therefore, it is suitable to use our sampling method to keep the most diverse samples to benefit the model performance on the coreset.

\newpage

\begin{CJK*}{UTF8}{gbsn}
\begin{table*}[!t]
    \scriptsize
    \centering
    \begin{tabularx}{\linewidth}{>{\centering\arraybackslash}m{0.5cm} >{\raggedright\arraybackslash}m{\dimexpr\linewidth-6.2cm\relax} >
    {\centering\arraybackslash}m{1.2cm} >{\centering\arraybackslash}m{1.2cm} >
    {\centering\arraybackslash}m{1.2cm}}
    \toprule
    ID & \multicolumn{1}{c}{Example} & Frequency Distance & QuRating & Perplexity \\
    \midrule
    496 & Euro-Scandinavian media cheer Denmark v Sweden draw. Denmark and Sweden tie. & 1.009 (99.91\%) & -2.65 (18.21\%) &  162.14 (99.39\%) \\
    \midrule
    904 & Rumsfeld said the Pentagon's annual assessment of China's military capabilities shows China is spending more than its leaders acknowledge, expanding its missile capabilities and developing advanced military technology. China was increasing its military spending and buying large amounts of sophisticated weapons. & 0.994 (77.10\%) & -2.56 (20.44\%) & 25.21 (36.30\%) \\
    \midrule
    787 & Guggenheim Museum, officially Solomon R. Guggenheim Museum, was founded in 1939 as the Museum of Non-Objective Art. The Solomon R. Guggenheim Museum was opened in 1939. & 0.998 (88.83\%) & -3.18 (6.97\%) & 14.76 (8.39\%)  \\
    \bottomrule
    \end{tabularx}
    \caption{Comparison to QuRating and perplexity of GPT-2 in RTE.}
\end{table*}
\end{CJK*}

\begin{CJK*}{UTF8}{gbsn}
\begin{table*}[!t]
    \scriptsize
    \centering
    \begin{tabularx}{\linewidth}{>{\centering\arraybackslash}m{0.5cm} >{\raggedright\arraybackslash}m{\dimexpr\linewidth-6.2cm\relax} >
    {\centering\arraybackslash}m{1.2cm} >{\centering\arraybackslash}m{1.2cm} >
    {\centering\arraybackslash}m{1.2cm}}
    \toprule
    ID & \multicolumn{1}{c}{Example} & Frequency Distance & QuRating  & Perplexity \\
    \midrule
    1019 & Tonight a spokesman for Russia 's foreign ministry said the ministry may issue a statement on Thursday clarifying Russia 's position on cooperation with Iran 's nuclear-energy efforts . Tonight a spokesman for the Russian Foreign Ministry said it might issue a statement on Thursday clarifying Russia 's position on aiding Iran 's nuclear-energy efforts . & 0.993 (71.01\%) & -2.54 (21.15\%) & 13.58 (9.73\%) \\
    \midrule
    2857 & Mr. Soros branded Mr. Snow \'s policy shift a " mistake . " Soros criticised Snow \'s policy shift as a " mistake " . & 1.006 (99.97\%) & -1.10 (69.98\%) & 94.34 (98.11\%)\\
    \midrule
    2208 & AAA spokesman Jerry Cheske said prices may have affected some plans , but cheap hotel deals mitigated the effect . AAA spokesman Jerry Cheske said prices might have affected some plans , but cheap hotel deals made up for it . & 0.998 (92.44\%) & -2.81 (13.79\%)& 57.03 (91.27\%) \\
    \bottomrule
    \end{tabularx}
    \caption{Comparison to QuRating and perplexity of GPT-2 in MRPC.}
\end{table*}
\end{CJK*}

\begin{CJK*}{UTF8}{gbsn}
\begin{table*}[!t]
    \scriptsize
    \centering
    \begin{tabularx}{\linewidth}{>{\centering\arraybackslash}m{0.5cm} >{\raggedright\arraybackslash}m{\dimexpr\linewidth-6.2cm\relax} >
    {\centering\arraybackslash}m{1.2cm} >{\centering\arraybackslash}m{1.2cm} >
    {\centering\arraybackslash}m{1.2cm}}
    \toprule
    ID & \multicolumn{1}{c}{Example} & Frequency Distance & QuRating  & Perplexity \\
    \midrule
    145 & It is important for the more you to eat, the more careful to be. & 0.958 (0.01\%) & -0.57 (4.63\%) & 124.01 (41.37\%) \\
    \midrule
    3576 & Students intended to surprise the teacher. & 0.989 (36.01\%) & 1.06 (52.08\%) & 228.78 (61.19\%) \\
    \midrule
    2940 & Donna fixed a sandwich. & 1.007 (99.71\%) & 1.23 (60.09\%) & 918.57 (89.70\%) \\
    \bottomrule
    \end{tabularx}
    \caption{Comparison to QuRating and perplexity of GPT-2 in CoLA.}
\end{table*}
\end{CJK*}

\begin{CJK*}{UTF8}{gbsn}
\begin{table*}[!t]
    \scriptsize
    \centering
    \begin{tabularx}{\linewidth}{>{\centering\arraybackslash}m{0.5cm} >{\raggedright\arraybackslash}m{\dimexpr\linewidth-6.2cm\relax} >
    {\centering\arraybackslash}m{1.2cm} >{\centering\arraybackslash}m{1.2cm} >
    {\centering\arraybackslash}m{1.2cm}}
    \toprule
    ID & \multicolumn{1}{c}{Example} & Frequency Distance & QuRating  & Perplexity \\
    \midrule
    22585 & The woman hands the girl to someone. Someone and someone & 0.924 (0.01\%) & 0.94 (39.75\%) & 187.70 (83.06\%) \\
    \midrule
    54915 & A yellow cab speeds towards him, then skids to a halt. Someone & 1.008 (99.99\%) & 1.60 (55.71\%) & 89.56 (53.67\%) \\
    \midrule
    53012 & Someone slowly gets up, locks eyes with someone. Someone looks guilty, weakly shaking his head, it & 0.980 (38.96\%) & 2.79 (81.53\%) & 80.09 (47.81\%) \\
    \bottomrule
    \end{tabularx}
    \caption{Comparison to QuRating and perplexity of GPT-2 in SWAG.}
\end{table*}
\end{CJK*}

\begin{CJK*}{UTF8}{gbsn}
\begin{table*}[!t]
    \scriptsize
    \centering
    \begin{tabularx}{\linewidth}{>{\centering\arraybackslash}m{0.5cm} >{\raggedright\arraybackslash}m{\dimexpr\linewidth-6.2cm\relax} >
    {\centering\arraybackslash}m{1.2cm} >{\centering\arraybackslash}m{1.2cm} >
    {\centering\arraybackslash}m{1.2cm}}
    \toprule
    ID & \multicolumn{1}{c}{Example} & Frequency Distance & QuRating  & Perplexity \\
    \midrule
    32532 & d )  & 0.096 (0.01\%) & -1.75 (2.87\%) & 1648.69 (72.21\%) \\
    \midrule
    10397 & in jerking off in all its byzantine incarnations to bother pleasuring its audience  & 0.991 (30.17\%) & 0.40 (28.98\%) & 208.79 (25.30\%) \\
    \midrule
    2940 & Donna fixed a sandwich. & 1.007 (99.71\%) & 1.62 (56.28\%) & 20360.19 (95.74\%) \\
    \bottomrule
    \end{tabularx}
    \caption{Comparison to QuRating and perplexity of GPT-2 in SST-2.}
\end{table*}
\end{CJK*}

\begin{CJK*}{UTF8}{gbsn}
\begin{table*}[!t]
    \scriptsize
    \renewcommand{\arraystretch}{1.2}
    \centering
    \begin{tabular}{>{\arraybackslash}m{12cm}|>{\centering\arraybackslash}m{2cm}}
    \hline
    \multicolumn{1}{c|}{Sample} & FD \\
    \hline
    \multicolumn{2}{c}{Furthest samples to geometric median} \\
    \hline
    What does 基督徒 (pinyin: jīdū tú) mean? (pinyin: jīdū tú), literally "Christ follower. & 1.009 \\ \hline
    Who wrote Carmen? Georges Bizet's Carmen premiered 3 March 1875. & 1.009 \\\hline
    Which Tigranes successor composed Greek tragedies? Tigranes' successor Artavasdes II even composed Greek tragedies himself. & 1.009 \\\hline
    What mostly affects polarization? Reflections generally affect polarization. & 1.009 \\\hline
    What does Orthodoxy strongly condemn? Similarly, Orthodoxy strongly condemns intermarriage. & 1.009 \\\hline
    \multicolumn{2}{c}{Nearest samples to geometric median} \\\hline
    On which side of the war were the Chinese? The major Allied participants were the United States, the Republic of China, the United Kingdom (including the armed forces of British India, the Fiji Islands, Samoa, etc.), Australia, the Commonwealth of the Philippines, the Netherlands (as the possessor of the Dutch East Indies and the western part of New Guinea), New Zealand, and Canada, all of whom were members of the Pacific War Council. & 0.954 \\\hline
    What was the name of the Philippines nation? The major Allied participants were the United States, the Republic of China, the United Kingdom (including the armed forces of British India, the Fiji Islands, Samoa, etc.), Australia, the Commonwealth of the Philippines, the Netherlands (as the possessor of the Dutch East Indies and the western part of New Guinea), New Zealand, and Canada, all of whom were members of the Pacific War Council. & 0.956 \\\hline
    The Battle of the Chernaya took place in what year? The deployment of Italian troops to the Crimea, and the gallantry shown by them in the Battle of the Chernaya (16 August 1855) and in the siege of Sevastopol, allowed the Kingdom of Sardinia to be among the participants at the peace conference at the end of the war, where it could address the issue of the Risorgimento to other European powers. & 0.956 \\\hline
    When were the economic laws passed in Mexico City? The politics pursued by the administrations of heads of government in Mexico City since the second half of the 20th century have usually been more liberal than those of the rest of the country, whether with the support of the federal government—as was the case with the approval of several comprehensive environmental laws in the 1980s—or through laws recently approved by the Legislative Assembly. & 0.957 \\\hline
    What political leaning does Mexico City take? The politics pursued by the administrations of heads of government in Mexico City since the second half of the 20th century have usually been more liberal than those of the rest of the country, whether with the support of the federal government—as was the case with the approval of several comprehensive environmental laws in the 1980s—or through laws recently approved by the Legislative Assembly. & 0.958 \\\hline
    \end{tabular}
    \caption{Nearest samples to geometric median and furthest samples to geometric median on the QNLI dataset.}
    \label{tab:example}
\end{table*}
\end{CJK*}

\end{document}